\documentclass{article}

\usepackage[dblblindworkshop, final]{neurips_2025}

\usepackage{graphicx}
\usepackage[english]{babel}

\usepackage{xcolor}

\usepackage{pgfplots}
\usepackage{pgfplotstable}
\pgfplotsset{compat=newest}
\usepgfplotslibrary{groupplots,statistics,colorbrewer}
\usepackage{tikz}
\usetikzlibrary{plotmarks}

\pgfdeclareplotmark{baseline}{\pgfuseplotmark{*}}
\pgfdeclareplotmark{fine-tuned grpo}{\pgfuseplotmark{square}}
\pgfdeclareplotmark{fine-tuned drgrpo}{\pgfuseplotmark{star}}
\pgfdeclareplotmark{qat 8bit STE}{\pgfuseplotmark{diamond}}
\pgfdeclareplotmark{ptq 8bit bnb}{\pgfuseplotmark{triangle}}
\pgfdeclareplotmark{ptq 8bit awq}{\pgfuseplotmark{pentagon}}
\pgfdeclareplotmark{qat 4bit qlora}{\pgfuseplotmark{oplus}}
\pgfdeclareplotmark{ptq 4bit bnb}{\pgfuseplotmark{otimes}}
\pgfdeclareplotmark{ptq 4bit awq}{\pgfuseplotmark{halfcircle}}

\tikzset{%
        baseline,
        inner sep=2pt,
        minimum height=12pt,
        rounded corners=2pt  
    }
\newcommand{\code}[1]{\mbox{% added this percent
        \ttfamily
        \tikz \node[anchor=base,fill=black!12]{#1};% added this percent
    }}

\tikzset{%
        baseline,
        inner sep=2pt,
        minimum height=12pt
    }

\usepackage{amsmath}
\usepackage{amsfonts}
\usepackage{amssymb}

\usepackage{graphicx}
\usepackage{multirow}
\usepackage{tabularx}
\usepackage[colorlinks=true, allcolors=blue]{hyperref}
\usepackage{siunitx}
\usepackage[utf8]{inputenc} % allow utf-8 input
\usepackage[T1]{fontenc}    % use 8-bit T1 fonts
\usepackage{hyperref}       % hyperlinks
\usepackage{url}            % simple URL typesetting
\usepackage{booktabs}       % professional-quality tables
\usepackage{nicefrac}       % compact symbols for 1/2, etc.
\usepackage{microtype}      % microtypography
\usepackage{xcolor}         % colors
\usepackage{comment}        % comment environment

\workshoptitle{Efficient Reasoning}

\usepackage{amsmath,amsfonts,bm}
\usepackage{mathtools}

\def\eqref#1{equation~\ref{#1}}
\def\1{\bm{1}}

\DeclareMathAlphabet{\mathsfit}{\encodingdefault}{\sfdefault}{m}{sl}
\SetMathAlphabet{\mathsfit}{bold}{\encodingdefault}{\sfdefault}{bx}{n}

\makeatletter
\renewcommand{\paragraph}{%
  \@startsection{paragraph}{4}%
  {\z@}{0.25ex \@plus 0.25ex \@minus .5ex}{-1em}%
  {\normalfont\normalsize\bfseries}%
}
\makeatother

\title{
The Impact of Quantization on Large Reasoning Model Reinforcement Learning
}

\author{Medha Kumar \\ Pennsylvania State University\thanks{Work done while interning at d-Matrix} \\ University Park, PA \\ mkumar@psu.edu 
\And
Zifei Xu \\ d-Matrix \\ Santa Clara, CA \\ xuzifei@d-matrix.ai
\And
Xin Wang \\ d-Matrix \\ Santa Clara, CA \\ poincare.disk@gmail.com
\And
Tristan Webb \\ d-Matrix \\ Santa Clara, CA \\ twebb@d-matrix.ai}

\begin{document}
\maketitle

\begin{abstract}
Strong reasoning capabilities can now be achieved by large-scale reinforcement learning (RL) without any supervised fine-tuning.  
Although post-training quantization (PTQ) and quantization-aware training (QAT) are well studied in the context of fine-tuning, how quantization impacts RL in large reasoning models (LRMs) remains an open question. 
To answer this question, we conducted systematic experiments and discovered a significant gap in reasoning performance on mathematical benchmarks between post-RL quantized models and their quantization-aware RL optimized counterparts. 
Our findings suggest that quantization-aware RL training negatively impacted the learning process, whereas PTQ and QLoRA led to greater performance. 
\end{abstract}

\section{Introduction}
Large reasoning models are trained in data and algorithmic pipelines. Commonly, LLMs are first "pre-trained" on trillions of input tokens, to develop a strong general ability to model the distribution of the training data, as well as showing ``sparks\ldots of intellegence''~\citep{bubeck2023sparks}. Post pre-training, models undergo further fine tuning, and one popular technique pioneered by \citet{shao2024deepseekmath} is to apply reinforcement learning to have the LLMs solve problems in domains with verifiable rewards, such as math or programming.

Quantization is a widespread technique for improving LLM memory and compute efficiency. As of 2025, new LLM "base" models commonly are released in FP$16$ or BF$16$ precision. Quantization is often left to fine tuners or software framework maintainers downstream from a model's official release. The intersection of RL and highly specialized agentic AI may lead to situations where many different LRM agents are derived from the same full precision base model, but specialized to different tasks through RL, and at some point quantized for inference performance. We study the question: how do we perform quantization to ensure the best test-time memory/performance tradeoff?

Prior work from authors \citet{krishnan2019quarl} found that during distributed training, the quantized actors could save energy. We are not aware of any other work examining the effect of quantization of training large reasoning models.

\section{Methods}
Our main investigation was to evaluate the reasoning performance of LLM models under different quantization strategies, and to explore the trade-offs between strategies in practice. Our general setup is that we have been provided with a base LLM which will be fine tuned through reinforcement learning on a specialized downstream tasks (such as mathematics) to produce a LLM with enhanced reasoning, or in other words a large reasoning model (LRM). Broadly speaking, a practitioner can choose from quantization strategies that can be divided into a number of different categories: such as post training quantization\,(PTQ), quantization aware training\slash fine-tuning\,(QAT\slash QAFT), and low-rank adapter fine tuning techniques, such as QLoRA. After quantization to a selected precision, we evaluate the models on a test dataset. 
We have released our training and evaluation code online at \href{https://github.com/d-matrix-ai/rlquant}{\texttt{github.com/d-matrix-ai/rlquant}}.

\begin{figure}[h!]
    \centering
    \begin{tikzpicture}
    \begin{axis}[no markers,
                cycle list/Set1-5,
                line width=1pt, 
                xlabel={Training Step}, 
                ylabel={Reward (mean)},
                xmin=0,
                xmax=600,
                ymin=0.55,
                ymax=1.0,
                each nth point=1,
                legend style={font=\small},
                legend pos=south east,
                 ]
% Just test data in here for now, put real data in rl_training.csv
        \addplot table [solid, mark=none x=globalstep, y=grpo, col sep=comma] {./data/training_ema_8b.csv};
        \addlegendentry{GRPO}
        \addplot table [solid, mark=none x=globalstep, y=drgrpo, col sep=comma] {./data/training_ema_8b.csv};
        \addlegendentry{drGRPO}
       \addplot table [solid, mark=none x=globalstep, y=lora, col sep=comma] {./data/training_ema_8b.csv};
        \addlegendentry{QLoRA}
     \addplot table [solid, mark=none x=globalstep, y=ste, col sep=comma] {./data/training_ema_8b.csv};
        \addlegendentry{$8$-bit STE}

       % \addplot table [solid, mark=none, x=global_step, y=Qwen3-1.7B-ft-drgrpo-bnb4 - eval/reward, col sep=comma] {./data/test_rl_training.csv};
        \addlegendentry{second plot}
    \end{axis}
    \end{tikzpicture}
    \caption{Mean training reward observed during RL training of Qwen3-8B, windowed moving average (window size $=25$) shown.}
    \label{fig:grpo}
\end{figure}
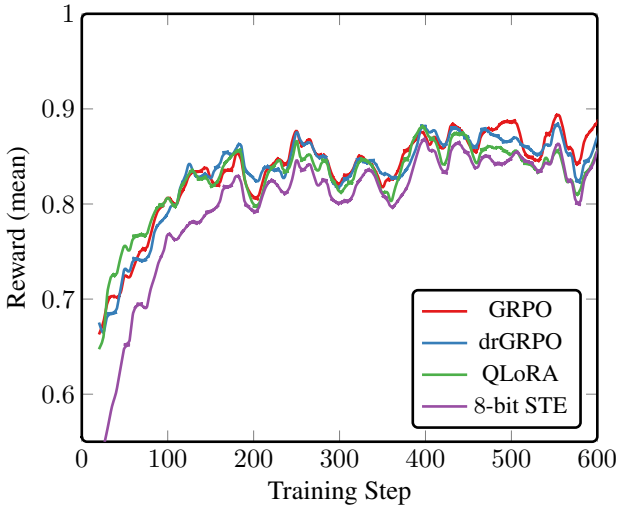
\subsection{Verifiable Reward RL training}
We utilized both the GRPO~\citep{shao2024deepseekmath} and drGRPO~\citep{liu2025understanding} algorithms to fine-tune base models from the Qwen3~\citep{yang2025qwen3} family of models on a variety of math benchmarks. We trained on the same MATH~\citep{hendrycks2021measuring} level 3-5 questions used by authors \citet{liu2025understanding}, and evaluated on questions from AIME2024, AMC, MATH500, Minerva Math and OlympiadBench on data also open sourced by the the same authors. Simulations were run using the GRPO and drGRPO training from the \code{TRL} library. Simulations were trained on \num[group-separator={,}]{10000} samples from the MATH dataset for $1$ epoch, with a learning rate of $10^{-6}$. In Fig.~\ref{fig:grpo} we show the training reward over the course of the run. Our reward function assigned a reward of $1$ for responses from the LLM that were mathematically correct, and furthermore added a reward of $0.1$ to responses that provided correct output formatting for the reward, which means rewards would be sampled from the set $\{0,0.1,1.1\}$. 

\subsubsection{QAFT with 8-bit STE}
One of the simplest methods to perform quantization aware fine tuning is before calculating activation to preform a Round to Nearest~(RTN) quantization to the weight and then perform straight through estimation~(STE) of the quantization gradients to avoid any non-differentiable sections of the graph. In our experiments, we quantized all of the linear layer weights (not activations) in the attention blocks to INT8.

\subsubsection{QLoRA}
Another quantization that is used during the RL training process is QLoRA~\citep{dettmers2023qlora}. This method is considered "parameter-efficient" since it introduces low-rank adapter matrices and during the RL process only the parameters in the smaller adapter matrices are updated while keeping the quantized weights frozen. QLoRA training was accomplished using the \code{PEFT} module from HuggingFace, and \code{bitsandbytes}(BnB) for quantization to NF4. QLoRA training used a learning rate of $10^{-4}$, a rank of $8$, and $\alpha=16$. The higher learning rate was required for the model to learn despite the quantization noise. QLoRA utilizes full numerical precision during model training, and the low rank adapter matrices can be merged into the base model resulting in a quantized model.

\subsection{PTQ via AWQ and bitsandbytes}
There are numerous PTQ techniques a modern practitioner could choose from to quantize a LRM after it has been fine-tuned. Beyond the other PTQ approaches we studied, there exist many other accessible approaches, such GPTQ~\citep{frantar2022gptq}, SpinQuant~\citep{liu2024spinquant}, GGUF~\citep{gerganov_llamacpp}, and many others. We chose two approaches that capture two different flavors of PTQ: data-free approaches, and those that use data to calibrate. With that, we selected the \code{bitsandbytes} and AWQ~\citep{lin2023awq} to produce PTQ models at both $8$ and $4$-bit precision. We applied PTQ directly to the same GRPO checkpoints we show evaluation results for in Table~\ref{tab:evaluation_math}. With \code{bitsandbytes} we specified a HuggingFace Quantization config to load the bit precision, and used the NF4 data type for $4$-bit quantization. We used the AWQ implementation from \code{llmcompressor}~\citep{llmcompressor2024}.

\section{Results}
Our main results are shown in Table~\ref{tab:evaluation_math}. We found that quantizing the network to $8$-bit precision through QAFT style STE training results in the greatest quantization error in networks larger than $0.6$B. We found the two PTQ techniques we examined performed well even at $4$-bits. Overall, using $4$-bit QLoRA during reinforcement learning training resulted in networks with the lowest quantization error in almost all cases. 
\begin{table}[t]
    \centering
    \begin{tabular}{|l|c|c|c|l|l|c|}
        \hline
        \multirow{2}{*}{Model} &
        \multicolumn{4}{c|}{Qwen} \\
         & $0.6$B & $1.7$B & $4$B & $8$B\\      \hline  
        (Base) &0.164  & 0.212 & 0.451 &0.473 \\
        \hline
        (GRPO) & \textbf{0.307} & \textbf{0.418}  & \textbf{0.555} & \textbf{0.594} \\
        (drGRPO) & 0.287  & 0.389& 0.541 & 0.584 \\
        \hline 
        (STE 8-bit)  & \textbf{0.242}  &0.325  & 0.443 & 0.496  \\
        (PTQ BnB 8-bit) &0.222 & \textbf{0.366} & \textbf{0.528} &0.579 \\
        (PTQ AWQ 8-bit) &0.22 & 0.364 & 0.526 & \textbf{0.583} \\
        \hline 
        (QLoRA 4-bit) & \textbf{0.24} & \textbf{0.382} &\textbf{ 0.554} & 0.556 \\
        (PTQ BnB 4-bit) & 0.223 &0.369 & 0.527 & \textbf{0.581} \\
        (PTQ AWQ 4-bit) &0.225 & 0.366 & 0.533 & 0.574 \\
        \hline
    \end{tabular}
    \vspace{10pt}
    \caption{Evaluation mean reward (rounded to 3 decimal places). (Base): full precision official Qwen3-Base models; (GRPO, drGRPO): full precision RL training; (STE): INT8 RTN, straight-through-estimator RL training; (PTQ): each method was performed on the full precision GRPO checkpoint evaluated above it.}
    \label{tab:evaluation_math} 
\end{table}

\subsection{Impact of completion length on model performance}
For the $0.6$B and $1.7$B models we set a completion length (for both training and evaluation) of $512$ tokens to obtain the evaluation scores shown in the results table. However, this completion length caused sub-optimal performance for the $4$B and the $8$B models. Table~\ref{tab:token_length} shows the mean evaluation reward at varying completion lengths for $4$B and $8$B. Increasing the completion length helped both models learn more from the same number of training steps.

\begin{table}[ht!]
    \centering
    \begin{tabular}{|l|c|l|l|c|}
        \hline
        \multirow{2}{*}{Model} &
        \multicolumn{2}{c|}{Qwen} \\
         & $4$B & $8$B\\      \hline  
        (GRPO @ 1024 tokens) &0.555 & 0.594\\
        \hline 
        (GRPO @ 512 tokens) &0.487 &0.540\\
        \hline
    \end{tabular}
    \vspace{10pt}
    \caption{Impact of token length on GRPO fine-tuned model performance. The same length is used in both training and evaluation.}
    \label{tab:token_length} 
\end{table}

\begin{comment}
\begin{figure}
    \centering
    \begin{tikzpicture}
    \begin{axis}[no markers,
                cycle list/Set1-5,
                line width=0.75pt, 
                xlabel={Training Step}, 
                ylabel={Reward},
                xmin=0,
                xmax=625,
                ymin=0,
                ymax=0.9,
                legend style={font=\small},
                legend pos=south east,
                 ]
        \addplot table [solid, mark=none, x=global_step, y=qwen3-8B-qat-ft-4bit-grpo-train/Reward/mean, col sep=comma] {./data/training_8B_curve.csv};
        \addlegendentry{4bit QAT run}
        \addplot table [solid, mark=none, x=global_step, y=Qwen3-8b-grpo-ft-train/rewards/Reward/mean, col sep=comma] {./data/training_8B_curve.csv};
        \addlegendentry{BF16 GRPO run}
    \end{axis}
    \end{tikzpicture}
    \caption{This figure shows the training curve of the Qwen3-8B-Base model for BF$16$ fine-tuning (blue) and QAT $4$-bit (red) fine-tuning}
    \label{fig:training_curve_8B}
\end{figure}
\end{comment}
\subsection{Evaluation score versus model size}

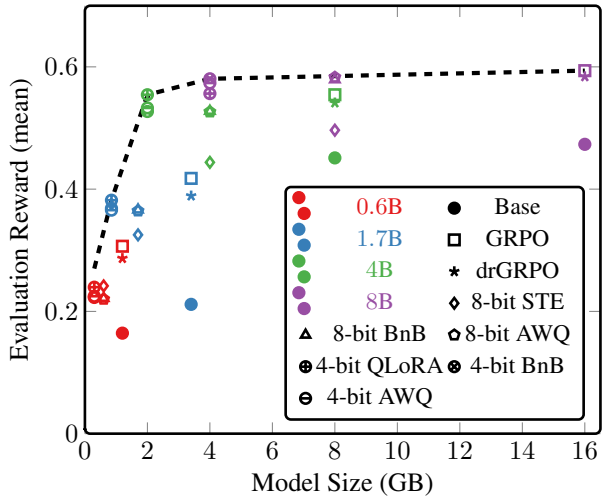
\begin{figure}[htb]
    \centering
    \begin{tikzpicture}
    \pgfplotsset{cycle list/Set1-4}
    \begin{axis}[
                scatter,
                scatter/@pre marker code/.append style={/tikz/mark=\Type},
                visualization depends on={value \thisrow{type} \as \Type},
                line width=1pt, 
                xlabel={Model Size (GB)}, 
                ylabel={Evaluation Reward (mean)},
                xmin=0,
                xmax=16.5,
                ymin=0,
                ymax=0.7,
                legend style={font=\small},
                legend pos=south east,
                legend columns = 2,
                legend entries={[index of colormap=0 of Set1-4]$0.6$B,Base,
                [index of colormap=1 of Set1-4]$1.7$B,GRPO, 
                [index of colormap=2 of Set1-4]$4$B,drGRPO,
                [index of colormap=3 of Set1-4]$8$B, 8-bit STE, 
                8-bit BnB,8-bit AWQ,
                4-bit QLoRA,4-bit BnB,
                4-bit AWQ
                }
            ]

    \addplot[only marks,xbar legend,index of colormap=0 of Set1-4, scatter/@pre marker code/.code={%
            \pgfplotsset{mark=\Type} \scope[index of colormap=0 of Set1-4]%
        },
    ] table [x=Qwen3-0.6B-Base_size_GB, y=Qwen3-0.6B-Base_mean-reward, col sep=comma] {./data/perf-vs-size--updated.csv}; 
    \addlegendimage{only marks,mark=*}
    \addplot[only marks,xbar legend,index of colormap=1 of Set1-4, scatter/@pre marker code/.code={%
            \pgfplotsset{mark=\Type} \scope[index of colormap=1 of Set1-4]%
        },
    ] table [x=Qwen3-1.7B-Base_size_GB, y=Qwen3-1.7B-Base_mean-reward, col sep=comma] {./data/perf-vs-size--updated.csv};
   \addlegendimage{only marks, mark=square}
    \addplot[only marks, xbar legend,index of colormap=2 of Set1-4, scatter/@pre marker code/.code={%
            \pgfplotsset{mark=\Type} \scope[index of colormap=2 of Set1-4]%
       },
    ] table [x=Qwen3-4B-Base_size_GB, y=Qwen3-4B-Base_mean-reward, col sep=comma] {./data/perf-vs-size--updated.csv};
    \addlegendimage{only marks, mark=star}
    \addplot[only marks,xbar legend,index of colormap=3 of Set1-4, scatter/@pre marker code/.code={%
            \pgfplotsset{mark=\Type} \scope[index of colormap=3 of Set1-4]%
       },
    ] table [x=Qwen3-8B-Base_size_GB, y=Qwen3-8B-Base_mean-reward, col sep=comma] {./data/perf-vs-size--updated.csv};
      \addlegendimage{only marks, mark=diamond}
       \addlegendimage{only marks, mark=triangle}
       \addlegendimage{only marks, mark=pentagon}
\addlegendimage{only marks, mark=oplus}
\addlegendimage{only marks, mark=otimes}
\addlegendimage{only marks, mark=halfcircle}
% \addplot+[ultra thick, dashed, black] coordinates{
%             (0.3,0.27)
%             (0.85,.3819)
%             (2,0.5544)
%             (4,0.5805)
%             (16,0.5938)
% };
\addplot+[ultra thick, dashed, black] table[x=x, y=y] {
            x   y   type
            0.3 0.27 0
            0.85    0.3819 0
            2   0.5544 0
            4   0.5805  0
            16  0.5938 0
};

    \end{axis}
    \end{tikzpicture}
    \caption{Evaluation reward vs. model size across all the models that we evaluated. We show the optimum pareto frontier as a dashed line.}
    \label{fig:training_curve_8B}
\end{figure}

Figure~\ref{fig:training_curve_8B} shows a plot of the performance\slash{memory} trade-off for different models we evaluated, across sizes, RL training, and quantization.
Our analysis shows that quantization generally offers stronger performance than using smaller full precision networks. 

\section{Discussion}
Our results show a strong trend across different model sizes. Techniques such as QAFT have generally been a sample efficient method\citep{xu2024understanding} to quantize neural networks and maintain performance. However, techniques like reinforcement learning produce a unique challenge for the quantization of LLMs, in that a discrete rewards are sampled from LLM generated responses, and quantized models produce worse policies. Our results show that techniques that quantize models downstream from training such as PTQ and QLoRA result in models that reason better on downstream tasks and result in better performance\slash{memory} trade offs. We find these techniques very effective at preserving reasoning ability, even at $4$-bit precision. 

Our study should not be interpreted discouraging the use of QAT during large reasoning models pre-training. Rather, we show that a sudden shock of quantization during the reinforcement learning process is damaging to learning. If QAT/QAFT was initiated prior to reinforcement learning training, perhaps the model would have already adapted to the lower bit-precision and would have learned effectively. We leave this and the discovery of more efficient quantization techniques for large reasoning models to future work, and present this work as a guide for the modern practitioner.
\newpage
\bibliography{main}

\end{document}